\documentclass{article}

\PassOptionsToPackage{numbers, compress}{natbib}

\usepackage[preprint]{neurips_data_2023}

\usepackage{graphicx}
\usepackage{multirow}
\usepackage{bbm}
\usepackage{subcaption}

\usepackage{amsmath}
\newcommand{\etal}{\textit{et al.}}

\usepackage{xcolor,pifont}
\newcommand*\colourcheck[1]{%
  \expandafter\newcommand\csname #1check\endcsname{\textcolor{#1}{\ding{52}}}%
}
\usepackage{color, colortbl}
\definecolor{Gray}{gray}{0.9}
\colourcheck{blue}
\colourcheck{green}
\colourcheck{red}

\newcommand{\figref}[1]{Fig. \ref{#1}}
\newcommand{\tabref}[1]{Tab. \ref{#1}}

\usepackage[utf8]{inputenc} 
\usepackage[T1]{fontenc}    
\usepackage{hyperref}       
\usepackage{url}            
\usepackage{booktabs}       
\usepackage{amsfonts}       
\usepackage{nicefrac}       
\usepackage{microtype}      
\usepackage{xcolor}         

\title{Universal Noise Annotation: Unveiling the Impact of Noisy annotation on Object Detection}

\author{
    Kwangrok Ryoo\thanks{These authors contributed equally to this work}$^{*1}$, Yeonsik Jo$^{*1}$, Seungjun Lee$^{*1}$, Mira Kim$^{*2}$,\\
    \textbf{Ahra Jo$^1$, Seung Hwan Kim$^1$, Seungryong Kim$^2$, Soonyoung Lee$^1$}\\
  $^1$LG AI Research, $^2$Korea University\\
  \texttt{$\{$kwangrok.ryoo, yeonsik.jo, 
seungjun.lee, ahra.jo, sh.kim, soonyoung.lee$\}$}\\\texttt{@lgresearch.ai} \\
  \texttt{$\{$miramira227, seungryong\_kim$\}$@korea.ac.kr}
}

\begin{document}

\maketitle

\begin{figure}[ht!]
\begin{subfigure}{.24\textwidth}
  \centering
  \includegraphics[width=0.9\linewidth]{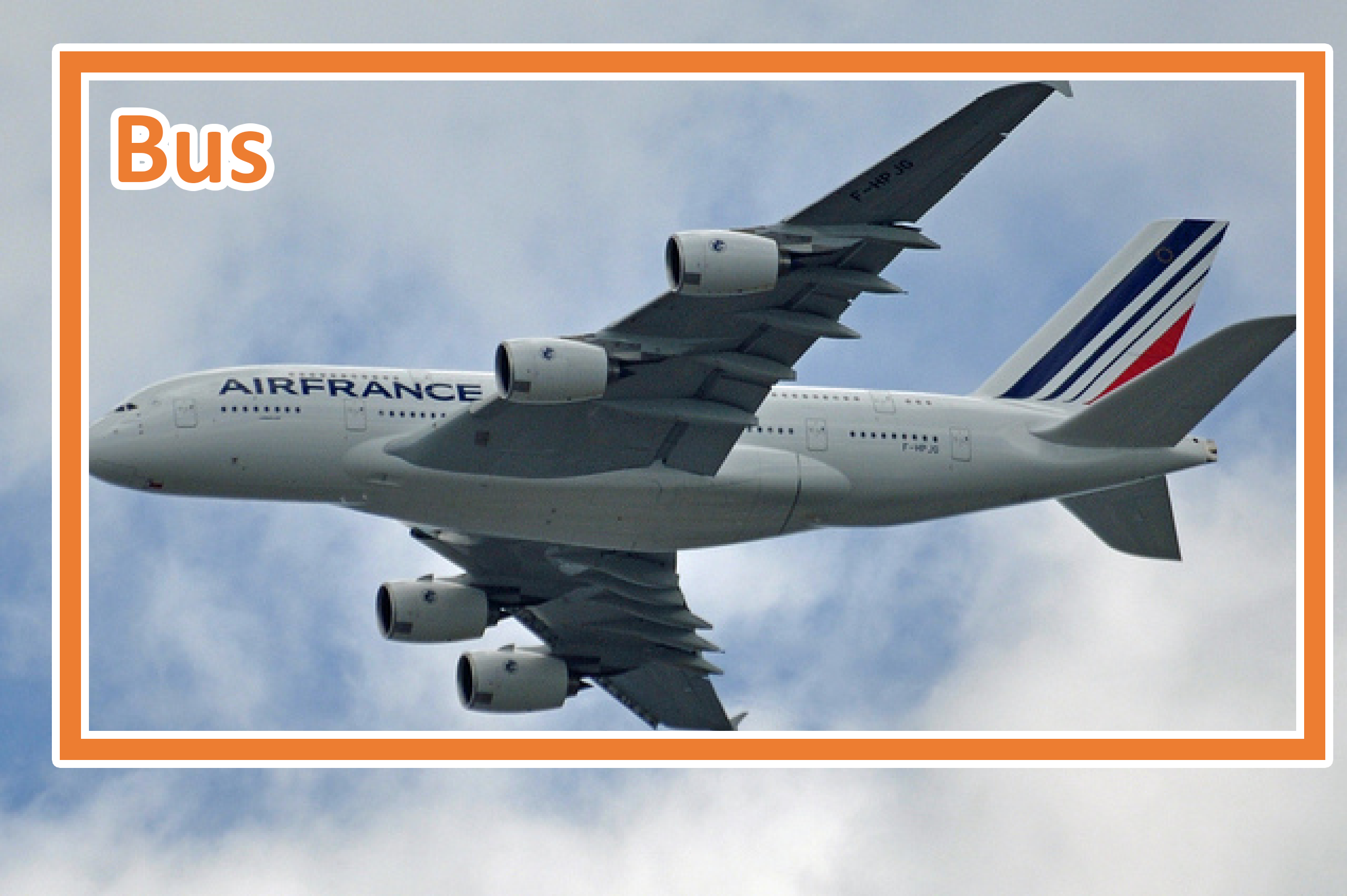}
  \caption{Categorization noise}
  \label{fig:noise1}
\end{subfigure}%
\begin{subfigure}{.24\textwidth}
  \centering
  \includegraphics[width=0.9\linewidth]{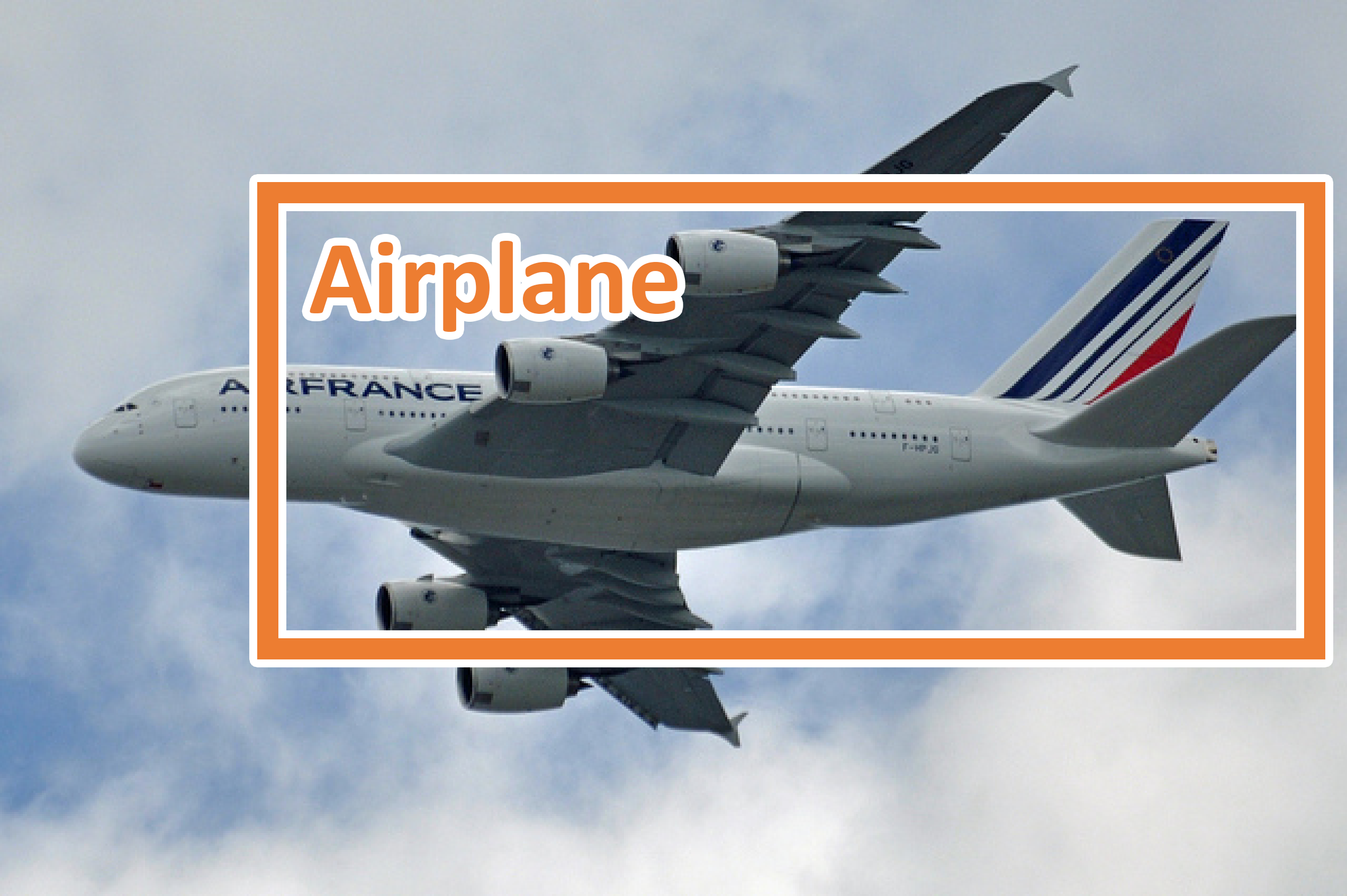}
  \caption{Localization noise}
  \label{fig:noise2}
\end{subfigure}%
\begin{subfigure}{.24\textwidth}
  \centering
  \includegraphics[width=0.9\linewidth]{imgs/missing.png}
  \caption{Missing annotation}
  \label{fig:noise3}
\end{subfigure}%
\begin{subfigure}{.24\textwidth}
  \centering
  \includegraphics[width=0.9\linewidth]{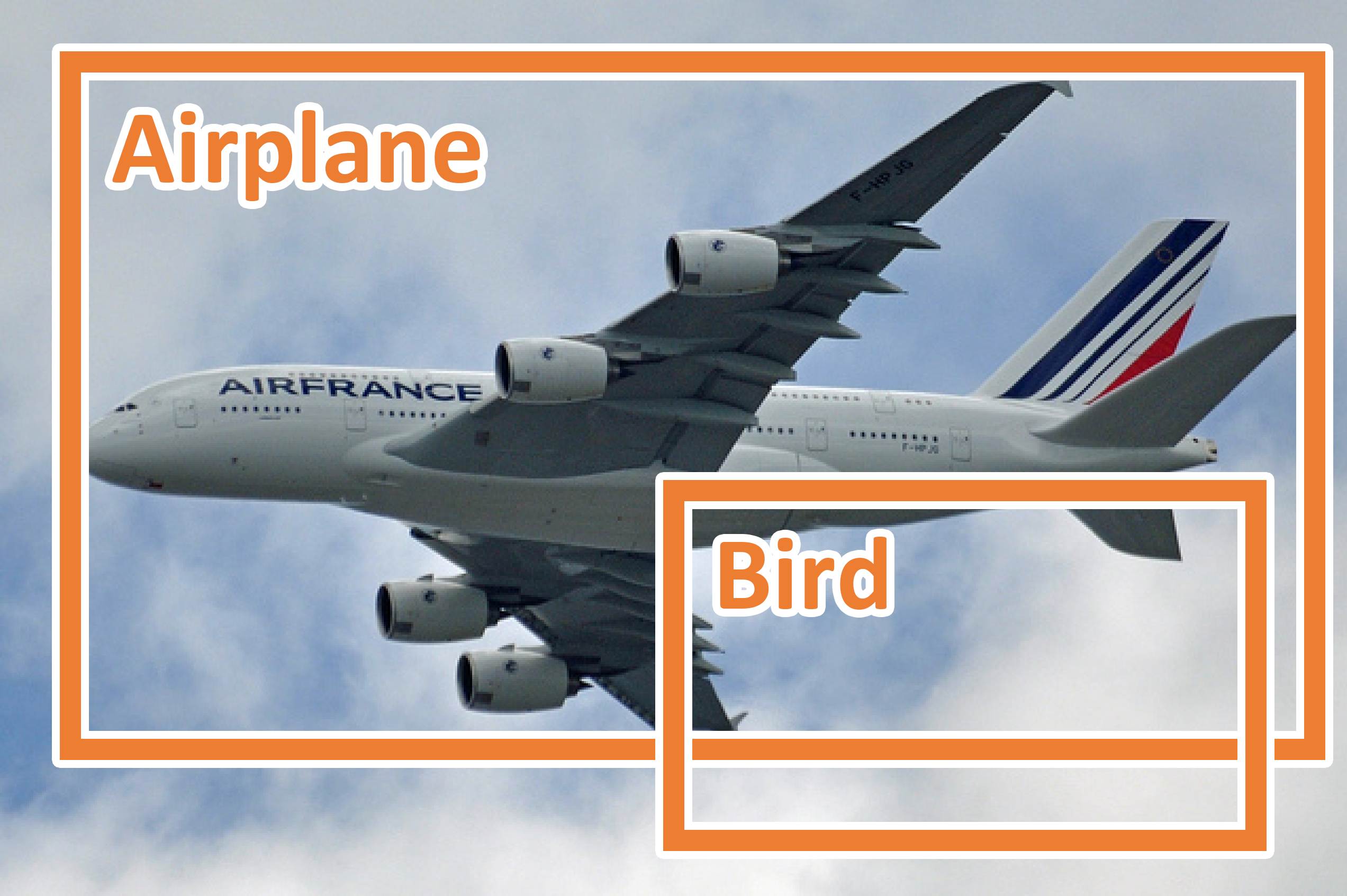}
  \caption{Bogus bounding box}
  \label{fig:noise4}
\end{subfigure}
\caption{\textbf{Visualization of possible noises in the object detection dataset:}  (a) category of bounding box can be mislabeled, (b) localization of bounding box can be wrong, (c) true object can be unlabeled in the dataset, and (d) a bogus bounding box of a random class at a random position can be present.}
\label{fig:noisetypes}
\end{figure}

\begin{abstract}
For object detection task with noisy labels, it is important to consider not only categorization noise, as in image classification, but also localization noise, missing annotations, and bogus bounding boxes. 
However, previous studies have only addressed certain types of noise (e.g., localization or categorization). 
In this paper, we propose Universal-Noise Annotation (UNA), a more practical setting that encompasses all types of noise that can occur in object detection, and analyze how UNA affects the performance of the detector. 
We analyzed the development direction of previous works of detection algorithms and examined the factors that impact the robustness of detection model learning method.
We open-source the code for injecting UNA into the dataset and all the training log and weight are also shared.
\end{abstract}

\section{Introduction}
\label{sec:intro}
Collecting well-annotated large-scale datasets has been essential for the success of deep learning~\cite{krizhevsky2012imagenet} in spite of its time-consuming and expensive process. 
It has become a natural practice to use crowd-sourcing or leverage machine-generated labels~\cite{su2012crowdsourcing}.
Yet these methods inevitably contain noisy data, which are mislabeled data, resulting in a significant decline in the performance of the model. 
Several attempts have been made to address this challenge, especially in image classification tasks~\cite{li2020towards, li2020dividemix, nishi2021augmentation}, which is commonly referred to as learning with noisy labels (LNL).

\begin{table}[ht!]
  \caption{\textbf{Comparison of noise type coverage.} For the first time, we consider four types of noise that are likely to occur in a detection task simultaneously.}
  \centering
  \resizebox{0.7\textwidth}{!}{
  \begin{tabular}{@{}ccccc@{}}
    \toprule
    & \multicolumn{4}{c}{Noise Type} \\
    \cmidrule(r){2-5}
    Methods & Categorization & Localization & Missing & Bogus \\
    \midrule
    CINJ~\cite{mao2021noisy}, MRNet~\cite{xu2021training} & \bluecheck & \bluecheck & \textbf{-} & \textbf{-} \\
    RefineDet~\cite{mao2021noisylocal}, OA-MIL~\cite{liu2022robust} & \textbf{-} & \bluecheck & \textbf{-} & \textbf{-} \\
    RSA~\cite{rambhatla2022sparsely}, HSL~\cite{xu2019missing} & \textbf{-} & \textbf{-} & \bluecheck & \textbf{-} \\
    \midrule
     \textbf{Ours (UNA)} &  \bluecheck &  \bluecheck &  \bluecheck &  \bluecheck\\
    \bottomrule
  \end{tabular}
  }
  \label{tab:Comparison}
\end{table}

However, unlike image classification task that primarily focuses on class labels, object detection task expands the concept of noise labels to include localization noise and even the presence or absence of objects. 
For example, \figref{fig:noisetypes} illustrates the four types of possible noise in crowd-sourcing or machine-generated labels; categorization noise~(\figref{fig:noise1}), localization noise~(\figref{fig:noise2}), missing annotation~(\figref{fig:noise3}), and bogus bounding box noise~(\figref{fig:noise4}). 
While the ideal choice to handle this problem is to undergo multiple steps of data label refinement, those processes can be highly expensive, challenging to guarantee a completely noise-free dataset.
Therefore, recent studies have emerged to analyze the impact of noise labels on the training of object detection models~\cite{mao2021noisy, xu2021training, mao2021noisylocal, liu2022robust, rambhatla2022sparsely, xu2019missing}.
However, we found that previous papers have only addressed a subset of possible label errors as shown in~\tabref{tab:Comparison}.
Since these annotation noises occur simultaneously, an overall investigation of all four noise types is necessary to gain a more thorough understanding of the challenges of learning from noise annotations and to develop robust training method for object detection tasks. 

For this end, we introduce \textbf{U}niversal-\textbf{N}oisy \textbf{A}nnotation (UNA) as a new benchmark to tackle the challenges posed by noise labels in object detection algorithms. 
Through a comprehensive analysis of previous methods that have aimed to improve detection performance, we have observed that the majority of these also demonstrate favorable results in the UNA setup. 
However, it is noteworthy that a subset of methods for utilizing loss values or auxiliary training may exhibit worse result than baseline. 

We conduct systemic analysis with a general toolkit named TIDE~\cite{bolya2020tide}.
TIDE is a framework to trace the sources of error in object detection. TIDE shows dAP ($\Delta$AP), which is the potential gain that can be obtained when improving each specific noise component in object detection.
With TIDE, we found the impact of different noise components and model design choices on the performance and robustness of object detection.

In summary, the key contributions of this paper are as follows:
\begin{itemize}
    \item We analyze the impact of noise labels on object detection training, considering factors such as including categorization noise, localization noise, missing annotations, and Bogus bounding boxes. 
    
    \item We propose a new benchmark named Unified Noise-Annotation (UNA) setup which addresses various types of noise in object detection at once.

    \item We present a comprehensive analysis of different detection models, including one-/two-stage, anchor-based/anchor-free, convolution/transformer backbone, and transformer-based models within the UNA setup.

\end{itemize}

Finally, our research addresses the challenge of limited code availability by ensuring the public accessibility of all code and implementations, promoting transparency and reproducibility in the field. 
To the best of our knowledge, UNA is the first publicly available benchmark for noise annotation settings in the field of object detection. 
The code, weight, experimental log and full paper for UNA can be found at \url{https://github.com/Ryoo72/UNA}.

\section{Related work}
In this section, we briefly review the previous studies and compare them with our approach.

\paragraph{Noisy labels datasets and benchmarks for classification task.}
Learning with noisy labels (LNL) trains models in the presence of incorrect labels, referred to as ``noisy'' labels, while correct labels are termed ``clean'' labels. 
The objective of LNL is to robustly train models despite the presence of noisy labels. 
The most widely used LNL benchmark is based on CIFAR datasets~\cite{krizhevsky2009learning}. 
There are two noise label settings: symmetric noise and asymmetric noise~\cite{tanaka2018joint,li2019learning}. 
Symmetric noise occurs when noise examples are randomly assigned to classes with equal probabilities. In contrast, asymmetric noise arises from randomly corrupting labels based on a pre-defined transition matrix. The corruption of examples in asymmetric noise is limited to similar classes, such as a dog being corrupted to a cat. 
However, not all instances of dog share similar features with cats. 
Therefore, to address this, CIFAR-IDN~\cite{chen2021beyond} was introduced, which sets the noise distribution on an instance-level rather than a class-level, providing a more realistic approach.
Our proposed benchmark, UNA, is designed to specifically address symmetric noise. This decision was made because incorporating asymmetric noise or instance dependency noise would require dataset-specific setups, limiting the generalizability of the benchmark to other datasets.

Subsequently, researchers proposed large-scale noisy labeled web-image databases such as Food-101N~\cite{bossard2014food,lee2018cleannet}, clothing1M~\cite{xiao2015learning}, and WebVision~\cite{li2017webvision} datasets to create real-world label noise. 
However, these datasets lack controllability as the noise labels are unknown. 
To address this issue, CIFAR-N~\cite{wei2021learning} dataset was introduced. 
CIFAR-N involved re-labeling the CIFAR dataset using Amazon Mechanical Turk to simulate human-biased noise. 
Unfortunately, creating new labels in object detection incurs significantly higher costs as well. Therefore, in this paper, we propose a new benchmark by categorizing and synthesizing noise that can occur during human annotation. 

\paragraph{Learning with noisy labels in object detection.}
Studies that involve learning with noisy labels task in object detection have started in recent years but only in small numbers~\cite{chadwick2019training, geiger2012we, li2020towards, mao2021noisylocal, xu2021training, liu2022robust}.
Chadwick~\etal~\cite{chadwick2019training} examined the effect of label noise on KITTI~\cite{geiger2012we} and proposed methods that mitigate the effect of noise using per-object co-teaching.
Li~\etal~\cite{li2020towards} proposed alternating two-step noise correction and model training that robustly train the model when localization noise and category noise are both present.
Mao~\etal~\cite{mao2021noisylocal} proposed a framework that mitigates localization noise by alternating refinement.
Xu~\etal~\cite{xu2021training} proposed a meta-learning based approach that utilizes few clean samples.
Liu~\etal~\cite{liu2022robust} leverages classification as a guidance signal for refining localization noise. 

Indeed, these methods suffer from the lack of direct comparability due to their use of individually tailored problem setups and the absence of publicly available source code. Further, most of methods tend to rely on specific structures or frameworks like FasterRCNN~\cite{ren2015faster}.
This creates a challenging situation for conducting meaningful comparisons. 
To tackle this problem, we introduce a unified benchmark, namely UNA. 
By leveraging the UNA benchmark, we conduct experiments that assess the robustness of detection algorithms in the presence of various types of noise. 

\section{Analysis and simulation for noise annotations in object detection}

\begin{figure}[ht!]
\begin{subfigure}{.24\textwidth}
  \centering
  \includegraphics[width=1\linewidth]{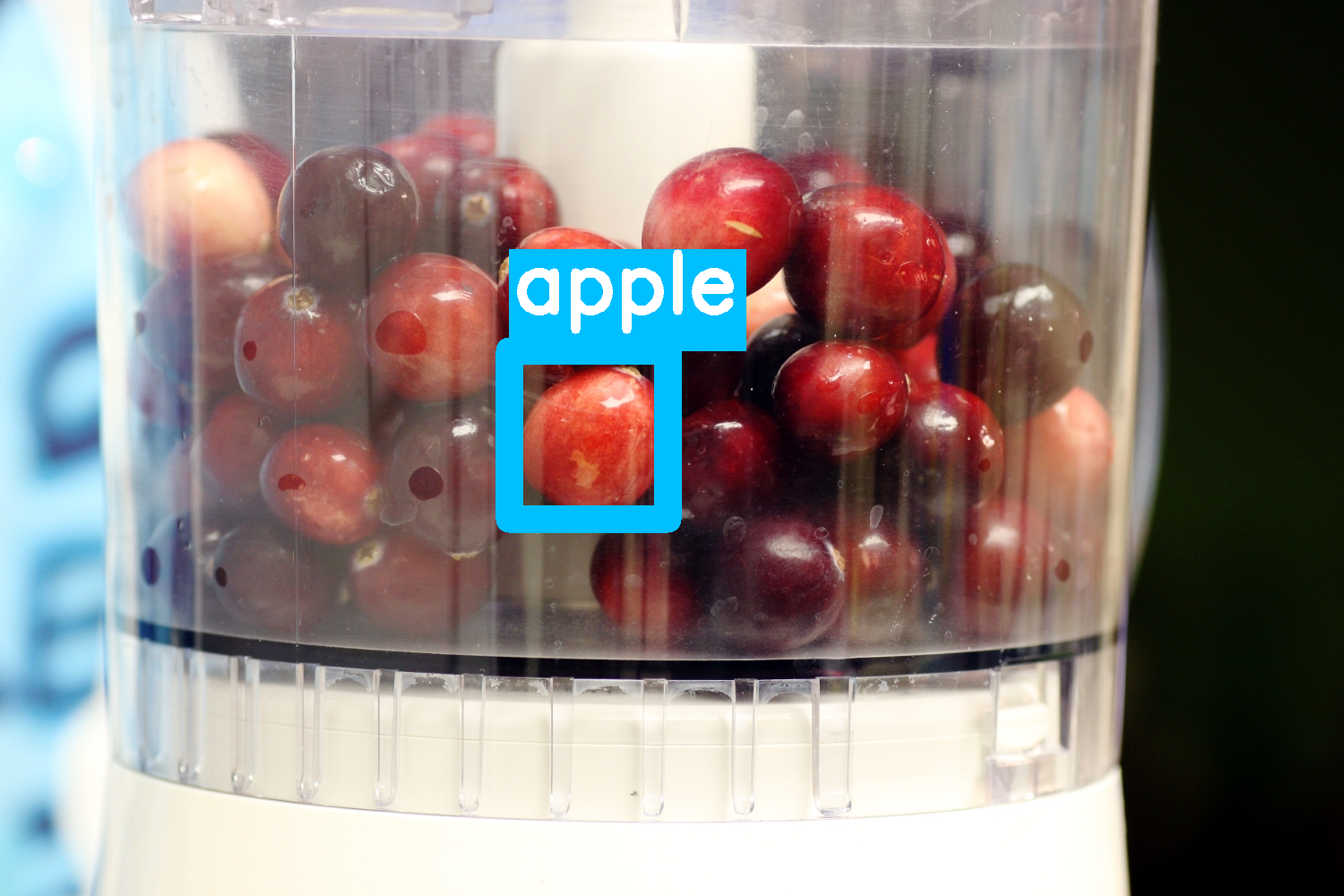}
  \caption{Categorization noise}
  \label{fig:OInoise1}
\end{subfigure}%
\hfill
\begin{subfigure}{.24\textwidth}
  \centering
  \includegraphics[width=1\linewidth]{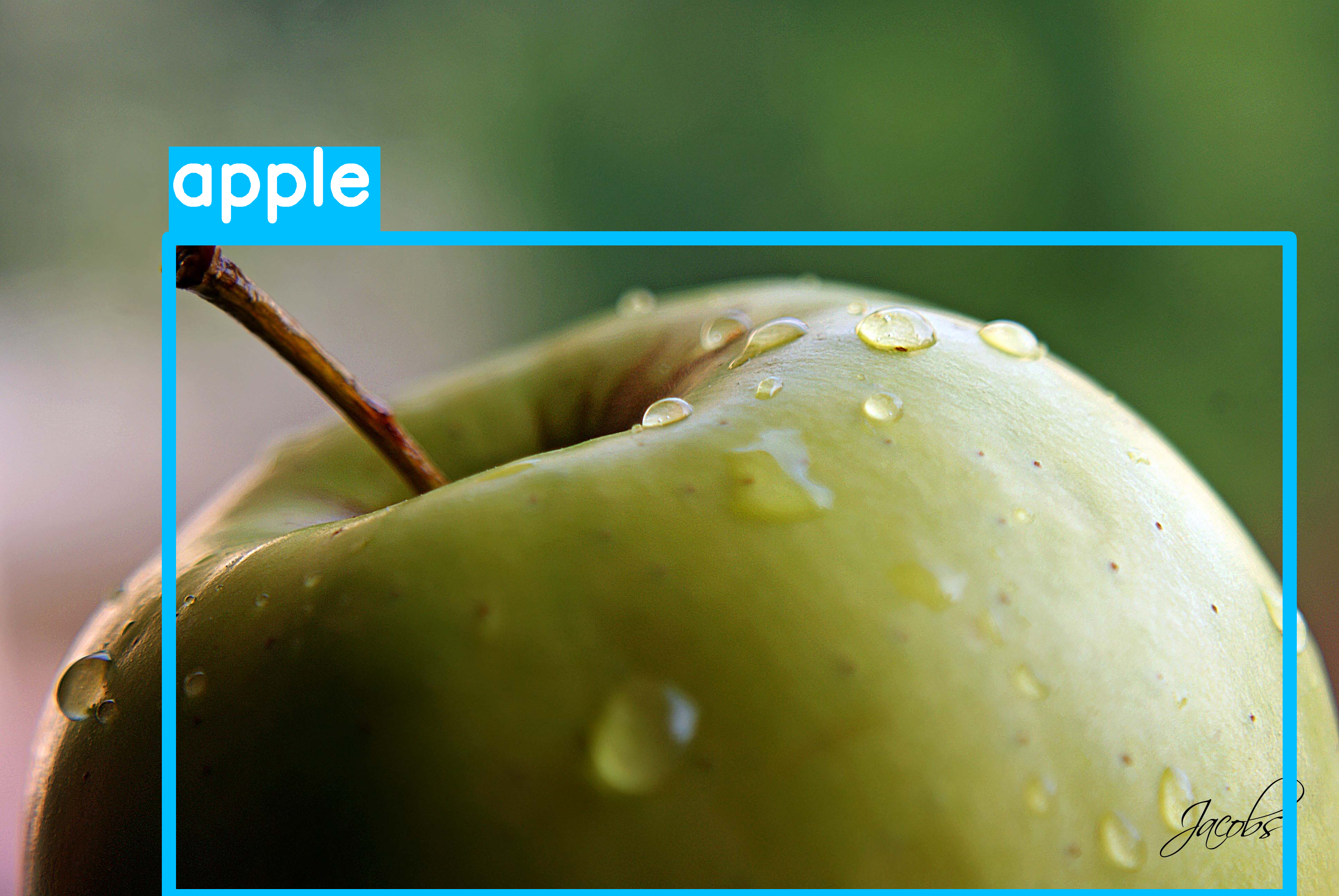}
  \caption{Localization noise}
  \label{fig:OInoise2}
\end{subfigure}%
\hfill
\begin{subfigure}{.24\textwidth}
  \centering
  \includegraphics[width=1\linewidth]{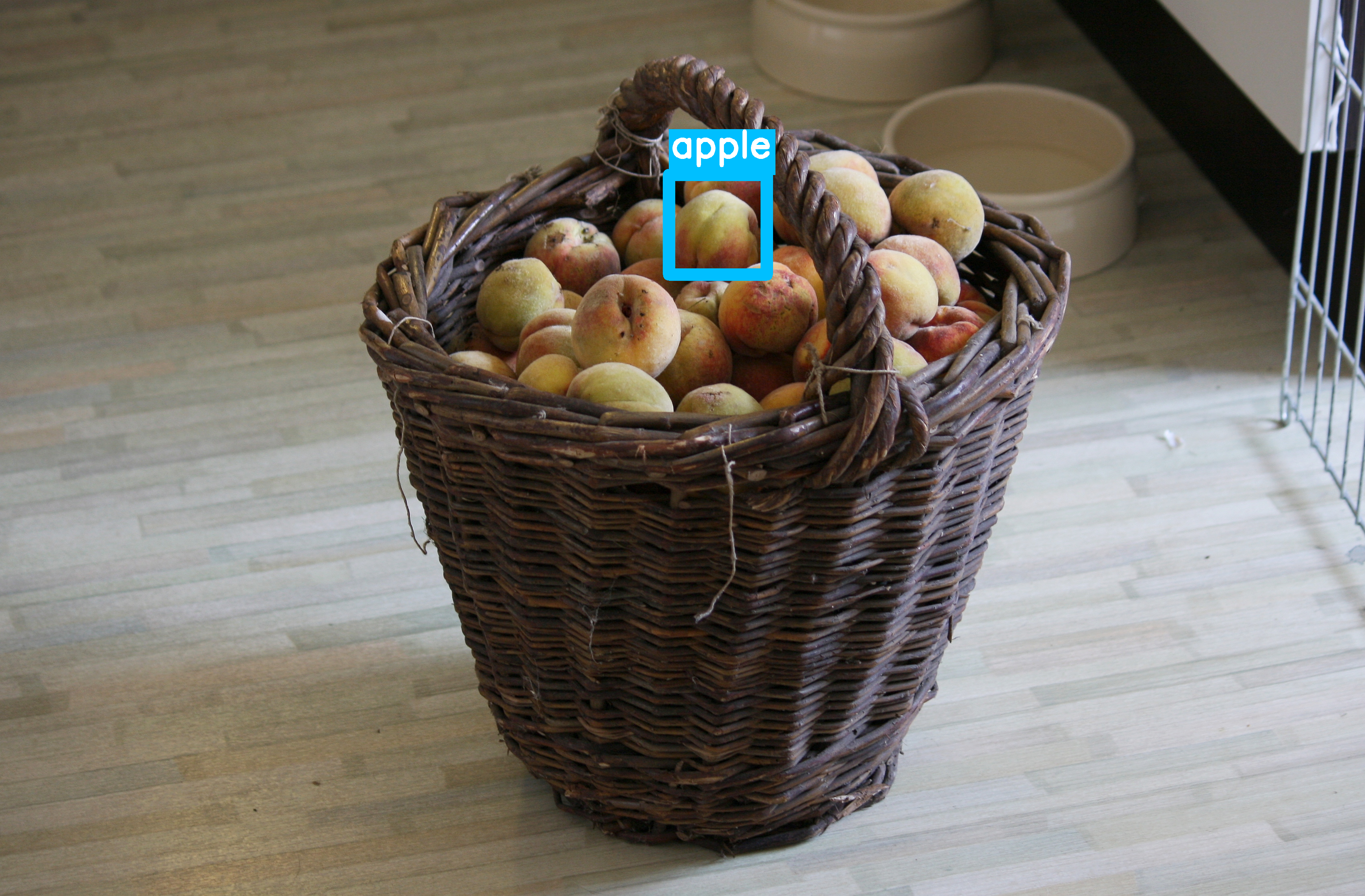}
  \caption{Missing annotation}
  \label{fig:OInoise3}
\end{subfigure}%
\hfill
\begin{subfigure}{.24\textwidth}
  \centering
  \includegraphics[width=1\linewidth]{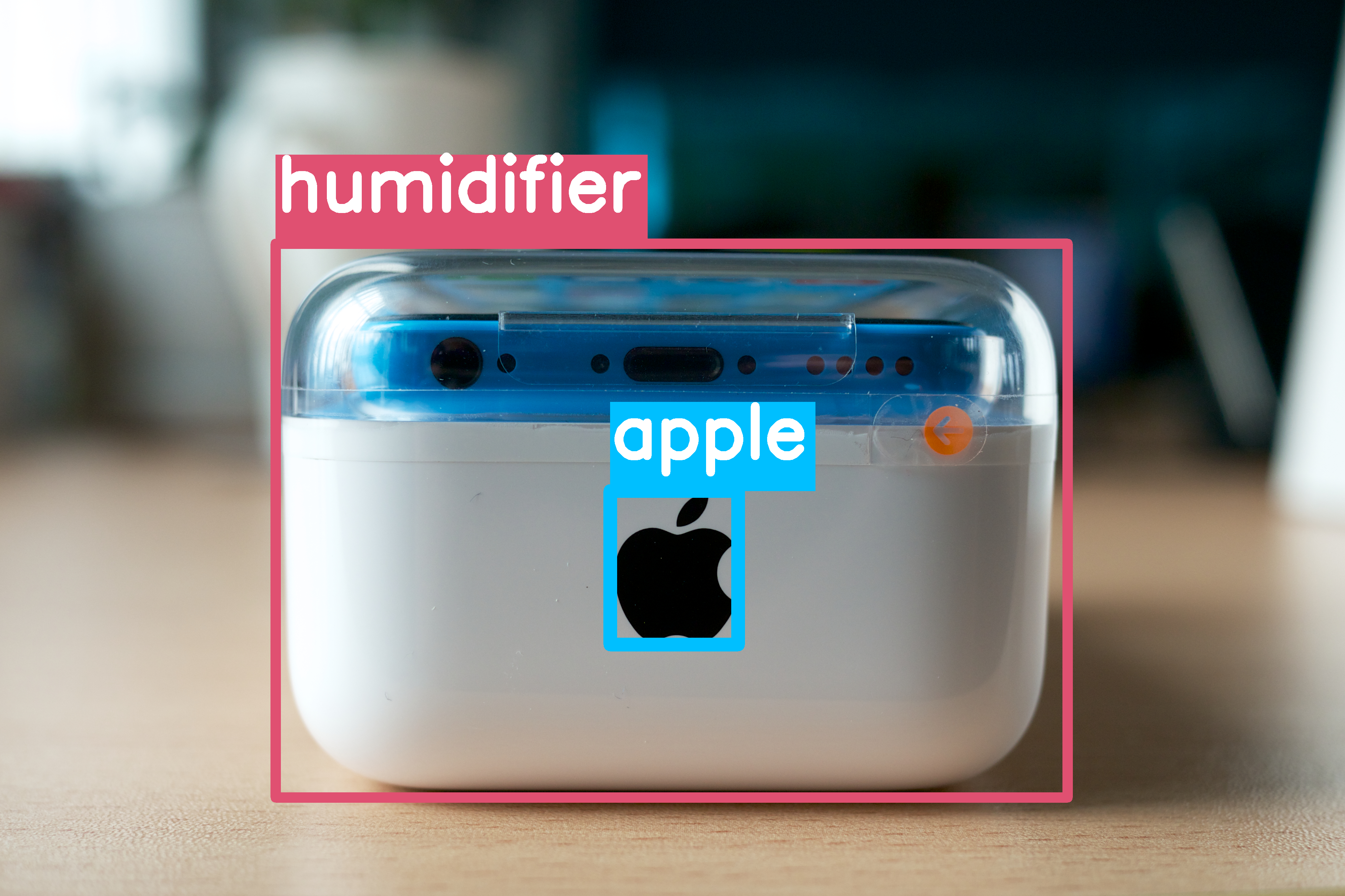}
  \caption{Bogus bounding box}
  \label{fig:OInoise4}
\end{subfigure}
\hfill
\begin{subfigure}{.49\textwidth}
  \centering
  \includegraphics[width=1\linewidth]{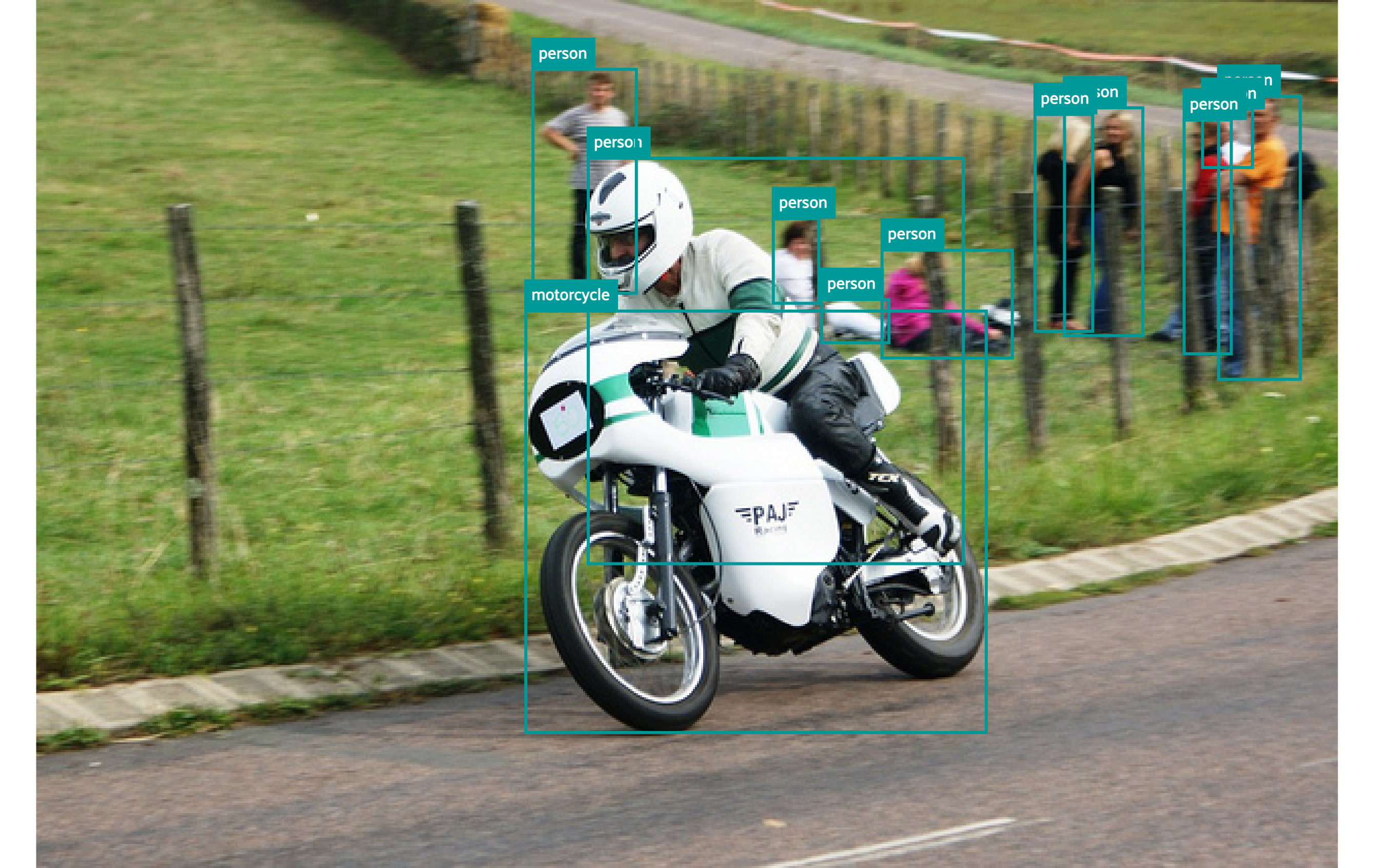}
  \caption{MS-COCO example}
  \label{fig:coco0}
\end{subfigure}
\hfill
\begin{subfigure}{.49\textwidth}
  \centering
  \includegraphics[width=1\linewidth]{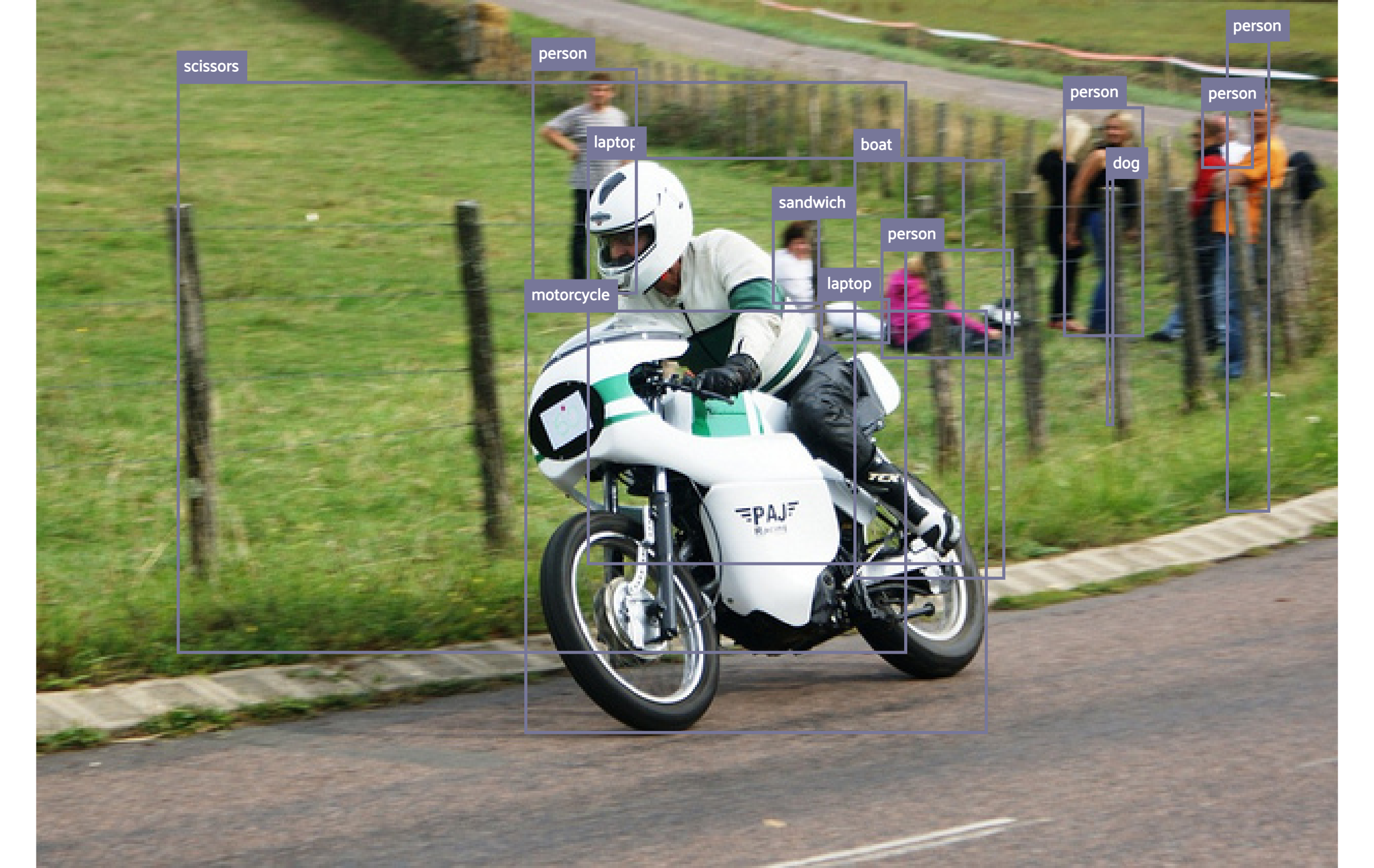}
  \caption{MS-COCO w/ UNA 20\%}
  \label{fig:coco20}
\end{subfigure}
\hfill
\caption{\textbf{Noisy annotations in OpenImages dataset~\cite{openimages} and UNA benchmark.} 
Real example on OpenImages dataset and proposed UNA on MS-COCO. Best viewed in color.
}
\label{fig:noiseindataset}
\end{figure}

In this section, we examine the different types of noise that can be present in a dataset. 
We categorize four types of noise as depicted in~\figref{fig:noisetypes}, each representing common types of noise that can be easily found in crowd-sourcing or machine-generated labels. 
\figref{fig:noiseindataset} represents specific noise types in real-world datasets, which will be described in more detail below and an instance of our new benchmark UNA. For more examples, please refer to our supplementary material.

\subsection{Object detection with noisy labels}
In object detection, an annotation set consists of categorical information and localization information such as bounding box coordinates.
In the context of object detection, noise labels deviate from the actual object's localization or categorization information. 
This misalignment can manifest as incorrect bounding box annotations, inaccurate class labels assigned to objects, or both.
We have categorized such cases into four different types.

\begin{itemize}
    \item \textbf{Categorization noise} occurs when the assigned category label for an object differs from its actual category. 
    Incorrect class tokens in object detection can lead to severe noise as it disrupts both the categorization and localization aspects.
    To ensure a dataset-agnostic setup, our focus is solely on symmetric noise, where categorical labels are randomly shifted with equal probability. By considering this type of noise, we aim to create a benchmark that is applicable across different datasets without relying on dataset-specific characteristics.
    \item \textbf{Localization noise} is the case where the location of an object $b_{i}$ is predicted incorrectly. 
    This type seems to be the most common in object detection tasks, as the criteria and subjective judgments of annotators can vary.
    Previous attempts~\cite{mao2021noisylocal,liu2022robust} to simulate localization noise in object detection often resulted in overly strong noise.
    Therefore, we introduce new noise generation scheme.
    For the details of algorithm, please refer to our supplementary material.

    \item \textbf{Missing annotation noise} refers to the absence or omission of annotations for certain objects. 
    Missing annotation noise is easily generated when objects are small or crowded. It is also known as sparse annotation noise~\cite{yoon2021semi,rambhatla2022sparsely,wang2021co} or positive-unlabeled noise~\cite{yang2020object} and there have been a relatively many numbers of studies compared to other noise. 
    While this type of noise can be connected to weakly-supervised learning, we primarily focus on scenarios where it occurs alongside other annotation noises within the annotation environment.
    
    \item \textbf{Bogus bounding box noise} refers to the presence of incorrect or misleading bounding boxes in the dataset. 
    It occurs when bounding boxes are mistakenly annotated for areas that do not contain any objects of interest. 
    These noise types can occur due to ambiguous definition of label or the use of machine-generated labels.
    For example, tagging the logo on the products of Apple Inc., such as `Macbook' or `iMac', to an apple or tagging the truck icon on a traffic sign to a truck are examples of bogus bounding box noise depending on the annotator and these cases are common in MS-COCO dataset~\cite{lin2014microsoft}. 
    Also, in medical image analysis domain, it can occur when a non-tumorous region is mistakenly tagged as a tumor with a bounding box.
\end{itemize}

\figref{fig:noiseindataset} compiles the noise types that can be found in the apple category of OpenImages dataset. 
In~\figref{fig:OInoise1}, a cranberry is labeled as apples. 
~\figref{fig:OInoise2} shows a case where the bounding box does not fully encompass the apple. 
In~\figref{fig:OInoise3}, only one apple is annotated while disregarding the others, and even this annotation has incorrect localization.
~\figref{fig:OInoise4} is an example of a bogus bounding box, where the logo is labeled as an apple. 
In this instance, there is also a categorization error as the phone package is mislabeled as a humidifier.

The aforementioned noise types do not occur individually but rather can occur simultaneously, with each type having an impact on the others. 
To address these interrelated noise types, we aim to develop a benchmark that can handle them collectively. 
In this context,  we proposed unified benchmark named \textbf{U}niversal-\textbf{N}oisy \textbf{A}nnotation (\textbf{UNA}).

\begin{itemize}
    \item \textbf{Universal Noise Annotation} is a unified benchmark, which combines categorization noise, localization noise, missing annotation noise, and bogus bounding box noise.
    UNA may consist of a simple combination of four types of noise, but there can also be synergistic effects among these noise types.
\end{itemize}

The specific algorithmic methods for generating each noises and UNA can be found in our supplementary material.
\figref{fig:coco20} is synthesized noisy data, where various types of noise have been added while preserving some degree of the actual labels on \figref{fig:coco0}. 

\subsection{Experiment and analysis}
\begin{table}[t]
\caption{\textbf{Experimental results on different noise types.} 
   The number in parentheses indicates the performance drop compared to clean label.
  }
  \centering
  \begin{tabular}{@{}l|c|cccc@{}}
    \toprule
    Noise type & 0\% & 5\% & 10\% & 15\% & 20\%\\
    \midrule
    Categorization & \multirow{5}{*}{37.4} 
                    & 36.5~(-0.9) & 36.0~(-1.4) & 35.7~(-1.7) & 35.0~(-2.4) \\
    Localization &  & 37.0~(-0.4) & 36.6~(-0.8) & 35.7~(-1.7) & 35.4~(-2.0) \\
    Missing &       & 37.1~(-0.3) & 36.8~(-0.6) & 36.4~(-1.0) & 36.1~(-1.3) \\
    Bogus &         & 37.2~(-0.2) & 36.9~(-0.5) & 36.6~(-0.8) & 36.5~(-0.9) \\
    \rowcolor{Gray}UNA && 35.7~(-1.7) & 33.1~(-4.3) & 29.9~(-7.5) & 26.6~(-10.8) \\
    \bottomrule
  \end{tabular}
  \vspace{1em}
  \label{tab:noiseperformance}
\end{table}

\begin{table}[t]
  \caption{\textbf{TIDE~($\Delta AP_{O}~(AP_{O})$) measure on noise ratio at 20$\%$.}
  We investigate the types of errors the model makes due to each noise component. Lower is better.
  }
  \centering
  \resizebox{1\textwidth}{!}{
  \begin{tabular}{@{}l|c|cccccc@{}}

    \toprule
    Noise type & AP$_{50}$ & Cls & Loc & Both & Dupe & Bkg & Miss  \\
    \midrule
    Clean      & 58.1 & 3.8~(62.0) & 7.0~(65.2) & 1.2~(59.3) & 0.3~(58.4) & 4.0~(62.1) & 8.0~(66.1)
 \\\hline
    Categorization & 55.7 & 4.8~(60.5) & 7.6~(63.3) & 1.3~(57.0) & 0.3~(56.0) & 3.9~(59.6) & 8.7~(64.4)
 \\
    Localization & 55.8 & 3.9~(59.8) & 7.2~(63.0) & 1.1~(56.9) & 0.3~(56.1) & 4.0~(59.8) & 8.5~(64.3)
 \\
    Missing & 57.0 & 4.3~(61.3) & 7.1~(64.1) & 1.3~(58.2) & 0.3~(57.3) & 3.9~(60.9) & 8.3~(65.3)
 \\
    Bogus & 57.2 & 4.1~(61.3) & 7.1~(64.3) & 1.2~(58.4) & 0.3~(57.5) & 4.0~(61.2) & 7.9~(65.1) 
\\
    \rowcolor{Gray}UNA & 43.3 & 4.0~(47.3) & 5.8~(49.1) & 0.9~(44.2) & 0.2~(43.5) & 7.2~(50.6) & 9.7~(53.0)
 \\
    \bottomrule
  \end{tabular}
  }
  \vspace{1em}
  \label{tab:TIDE}
\end{table}

In \tabref{tab:noiseperformance}, we evaluate the impact of the proposed noise types on object detection performance using MS-COCO benchmark. 
The experimental setup follows the configuration of MMdetection (Faster-RCNN-FPN-Resnet-50, 12 epochs). 
Among the four base types of noise, categorization noise had the most significant negative impact on performance, followed by localization noise, missing annotation noise, and bogus bounding box noise.
Categorization noise may work as adversarial noise that disrupts the learning process of other clean label, while other noise relatively has less impact on other samples. 
In our research, we have observed that the confidence score of object detection tends to decrease in the presence of noise labels. 
However, it is important to note that this decrease in confidence score may not be fully reflected in the mean Average Precision (mAP) metric, which is based on area under curve of precision-recall curves.

The observed significant performance degradation in the UNA dataset compared to other noise setups can be attributed to the fact that UNA has a higher level of noise because it include all kind of noise with same ratio.
Interestingly, the cumulative performance drop as each noise is added at a 5$\%$ noise level~($0.9+0.3+0.4+0.2 = 1.8$) is almost the same as the performance decrease observed in the UNA setup~($\sim1.7$). 
However, as the noise level approaches 20$\%$, the performance drop of UNA~($\sim10.8$) becomes even greater than the sum of the performance drops from the other four noise types ($2.4 + 2.0 + 1.4 + 0.9=6.7$).
This suggests that at lower noise levels, the four noise types operate independently, while at higher noise levels, they start to overlap and exhibit synergistic effects. 
Here, we argue that UNA serves as a valuable benchmark, which is not only summary of pre-existing noise label but also additional challenge to develop robust training method for object detection.

\subsection{The impact on the detector by each noise component}

In this section, we analyze the impact of each noise component (e.g., categorization noise, localization noise) on the model. To do so, we trained FasterRCNN using a 20\% noise ratio settings for each component and examined the types of errors the model made and used TIDE~\cite{bolya2020tide} to assess the model's errors as shown in~\tabref{tab:TIDE}.
TIDE is built upon the concept of delta average precision~($\Delta AP$), which represents the potential additional improvement achievable by fixing a specific component from the AP${50}$ baseline.
The fixed AP is denoted as oracle AP, AP$_{O}$, representing the ideal performance if selected component was perfect.
There are six components: Cls: classification error; Loc: localization error; Both: classification and localization error; Dupe: duplicated detection (another bounding box matching the Ground Truth (GT)); Bkg: detected background as foreground; and Miss: no GT to match prediction.

Through the analysis of the distribution of $\Delta AP$, we can gain insights into which aspects the model finds challenging. 
Interestingly, when using the basic four types of noise, we observe that the distribution of $\Delta AP$ does not change significantly. 
However, when applying the UNA setup, we notice a significant change in the distribution. This demonstrates that the situation can vary drastically between handling individual types of noise and dealing with all types of noise together.

For the visualizations and more detailed interpretations, please refer to Section 12 in the supplementary materials.

\section{Analyzing key contributions to robust detection learning with UNA}

In this section, we aim to investigate the influence of annotation noise on the development of detection algorithms using UNA benchmark. 
Our study focuses on analyzing how the advancement in detection model and feature extractor components, across different time periods, affects their ability to handle noise label training.
Note that we used 8 Tesla V100 GPUs for most of the experiments, and some were conducted using 8 GeForce RTX 3090 GPUs.

\subsection{Model design analysis} 

To thoroughly analyze the impact of the UNA setup on existing research, we select six baselines; FasterRCNN~\cite{ren2015faster}, OHEM~\cite{shrivastava2016training}, FCOS~\cite{tian2019fcos}, ATSS~\cite{zhang2020bridging}, DINO~\cite{zhang2022dino}.
All models follow the basic setup of the MMDetection framework~(FPN-Resnet-50, 12 epochs). 
We select the models with diverse characteristics to thoroughly analyze the factors that have had significant impacts on model design in the field of object detection.
~\tabref{tab:UNA} and~\tabref{tab:TIDE-algorithm} are the experimental results. For a more detailed analysis using LRP~\cite{oksuz2018localization}, please refer to Section 13 in the supplementary materials.

\paragraph{Hard example mining}
In the early stages of object detection, a key focus was on addressing the class imbalance issue between the background and foreground. To tackle this, several methods were developed to select background (negative samples). OHEM~\cite{shrivastava2016training} further improves performance by mining hard examples not only from the negative samples but also across a multi-class setting. However, OHEM, which uses high-loss samples for training, can mistakenly select wrong sample since the loss is calculated from noise label.
In clean data, OHEM surpasses the baseline FasterRCNN in terms of performance. However, when noise labels are introduced, the model's performance appears to deteriorate. This degradation is particularly evident in high value of Bkg and Miss on TIDE measure, which indicate an increased tendency of the model to confuse between background and foreground classes.

\paragraph{One-/Two-stage model}    
Next, we consider the comparison of one-/two-stage models, which have been widely utilized due to their respective trade-offs between performance and speed.
Two-stage model~(FasterRCNN~\cite{ren2015faster}) shows better performance on clean label setup, but in UNA setup, one-stage models~\cite{lin2017focal}(RetinaNet~\cite{lin2017focal}, FCOS~\cite{tian2019fcos}) show better result.
Since Region Proposal Network(RPN) in two-stage model is also trained using labels, two-stage model seems to be more sensitive to the impact of noise labels. 

\begin{table*}[t]
\caption{\textbf{Experimental results for various types of detectors on the UNA setting.} We experiment with representative models of one-/two-stage, anchor-based/anchor-free, and transformer head.}
  \centering
  \resizebox{1\linewidth}{!}{
  \begin{tabular}{@{}l|ccccccccccccccc@{}}
    \toprule
    &\multicolumn{5}{c}{$\mathrm{AP}$}&\multicolumn{5}{c}{$\mathrm{AP}_{50}$}&\multicolumn{5}{c}{$\mathrm{AP}_{75}$} \\\cmidrule(lr){2-6}\cmidrule(lr){7-11}\cmidrule(lr){12-16}
    Detector & 0\% & 10\% & 20\% & 30\% & 40\% & 0\% & 10\% & 20\% & 30\% & 40\% & 0\% & 10\% & 20\% & 30\% & 40\% \\
    \hline
    FasterRCNN~\cite{ren2015faster} & 37.4 & 33.1 & 26.6 & 16.2 & 1.0 & 58.1 & 52.8 & 43.3 & 27.1 & 1.5 & 40.4 & 35.9 & 28.4 & 17.1 & 1.0 \\
    OHEM~\cite{shrivastava2016training} &  37.7 & 32.9 & 25.3 & 13.1 & 0.3 & 58.5 & 51.9 & 40.3 & 21.2 & 0.4 & 41.2 & 34.6 & 27.7 & 14.4 & 0.3 \\
    RetinaNet~\cite{lin2017focal} & 36.5 & 32.4 & 27.6 & 17.1 & 1.8 & 55.4 & 50.9 & 44.6 & 29.1 & 3.5 & 39.1 & 34.4 & 29.1 & 17.6 & 1.6 \\
    FCOS~\cite{tian2019fcos} & 36.5 & 33.2 & 29.7 & 23.4 & 9.0 & 56.0 & 52.2 & 47.7 & 39.3 & 16.8 & 38.8 & 35.2 & 31.3 & 24.3 & 8.6 \\
    ATSS~\cite{zhang2020bridging} & 39.4 & 36.1 & 32.4 & 26.2 & 11.8 & 57.5 & 53.9 & 49.5 & 41.5 & 20.6 & 42.6 & 39.0 & 34.8 & 27.6 & 12.0 \\
    DINO~\cite{zhang2022dino} & 49.0 & 43.6 & 36.5 & 29.7 & 15.5 & 66.6 & 60.6 & 52.7 & 43.7 & 24.7 & 53.5 & 47.0 & 39.7 & 31.5 & 24.7 \\
    \bottomrule
  \end{tabular}
  }
  \label{tab:UNA}
\end{table*}

\begin{table}[t]
\caption{\textbf{TIDE~($\Delta AP_{O}~(AP_{O})$) on noise ratio at 20$\%$.}
  We investigate the types of errors the model makes due to each noise component. Lower is better. For number in parentheses, higher is better.
  }
  \centering
  \resizebox{1\textwidth}{!}{
  \begin{tabular}{@{}l|c|cccccc@{}}
    
    \toprule
    Model & AP$_{50}$ & Cls & Loc & Both & Dupe & Bkg & Miss  \\
    \midrule
    FasterRCNN~\cite{ren2015faster}      & 43.3 & 4.0~(47.3) & 5.8~(49.1) & 0.9~(44.2) & 0.2~(43.5) & 7.2~(50.6) & 9.7~(53.0)
 \\
    OHEM~\cite{shrivastava2016training} & 40.3 & 3.8~(44.0) & 5.2~(45.5) & 0.7~(40.9) & 0.1~(40.4) & 8.4~(48.7) & 11.7~(52.0)
 \\
    RetinaNet~\cite{lin2017focal} & 44.6 & 6.5~(51.0) & 6.4~(51.0) & 1.0~(45.6) & 0.3~(44.9) & 3.0~(47.6) & 7.4~(52.0)
 \\
    FCOS~\cite{tian2019fcos} & 47.7 & 5.6~(53.3) & 7.8~(55.5) & 1.1~(48.8) & 0.1~(47.8) & 4.0~(51.7) & 6.7~(54.4)
 \\
    ATSS~\cite{zhang2020bridging} & 49.5 & 5.3~(54.8) & 6.6~(56.0) & 1.2~(50.7) & 0.7~(50.2) & 3.6~(53.1) & 6.4~(55.8)
\\
    DINO~\cite{zhang2022dino} & 52.7 & 3.8~(56.5) & 7.0~(59.7) & 1.1~(53.8) & 1.4~(54.0) & 3.9~(56.6) & 5.0~(57.7)
\\
    \bottomrule
  \end{tabular}
  }
  \label{tab:TIDE-algorithm}
\end{table}

\paragraph{Anchor-based/Anchor-free}
While anchor-based models~(FasterRCNN, RetinaNet) were initially prevalent in detection models, the emergence of feature pyramid network(FPN)~\cite{lin2017feature} and focal loss~\cite{lin2017focal} paved the way for the development of anchor-free models~(FCOS). 
Here, the anchor based method~(RetinaNet) and anchor-free~(FCOS) method shows similar result on clean label but anchor based method shows more sensitive result on UNA setup.
Modifying the tuning of anchors may lead to different trends but optimal setup also vary.
With \tabref{tab:TIDE-algorithm}, anchor-based methods tend to exhibit greater robustness compared to anchor-free methods on localization, while are relatively weaker in terms of Cls compared to anchor-free methods.
ATSS~\cite{zhang2020bridging} combines the strengths of anchor-based and anchor-free methods by proposing a novel training data sampling method. It aims to address the limitations of both approaches and achieve improved performance in both general scenarios and the UNA setup.

\paragraph{Transformer Head}
The development of detection heads has led to the transformation of architectures into single networks like transformers. 
Early approaches like DETR~\cite{carion2020end} showed instability and performance similar to FasterRCNN, but advancements like Deformable DETR~\cite{zhu2020deformable}, Conditional DETR~\cite{meng2021conditional}, and DAB-DETR~\cite{liu2022dab} achieved stability and improved performance. 
Here, we select DINO~\cite{zhang2022dino} because it trains for 12 epochs, consistent with the other models we used for comparison. This contrasts with many other transformers that require significantly more epochs, making DINO a fairer comparison.
The results on more DETR-variants can be found in supplementary material.
DINO outperforms other methods in the baseline, demonstrating robust performance even in all ratio of UNA, as it is not affected by dataset-dependent hyper-parameters.

Throughout our examination of early and recent detection papers, we have observed a consistent trend in the development of detection algorithms aligning with increased robustness to noise annotations.

\subsection{Backbone analysis}
\label{sec:Backbone}

\begin{table*}[t]
  \caption{\textbf{Experimental results for various types of backbone on the UNA setting.} We experiment with differenct kind of backbone; Resnet-50, Deformable Convolution v2~\cite{zhu2019deformable}, Swin Transformer~\cite{liu2021swin} and ConvNeXt~\cite{liu2022convnet}. All detectors were based on FasterRCNN~\cite{he2016deep}.}
  \centering
  \resizebox{1\linewidth}{!}{
  \begin{tabular}{@{}l|ccccccccccccccc@{}}
    \toprule
    &\multicolumn{5}{c}{$\mathrm{mAP}$}&\multicolumn{5}{c}{$\mathrm{AP}_{50}$}&\multicolumn{5}{c}{$\mathrm{AP}_{75}$} \\\cmidrule(lr){2-6}\cmidrule(lr){7-11}\cmidrule(lr){12-16}
    Backbone & 0\% & 5\% & 10\% & 15\% & 20\% & 0\% & 5\% & 10\% & 15\% & 20\% & 0\% & 5\% & 10\% & 15\% & 20\% \\
    \hline
    ResNet~\cite{he2016deep} & 37.4 & 35.7 & 33.1 & 29.9 & 26.6 & 58.1 & 56.2 & 52.8 & 48.4 & 43.3 & 40.4 & 38.6 & 35.9 & 32.2 & 28.4 \\
    DCNv2~\cite{zhang2020bridging} &  41.4 & 38.8 & 36.6 & 31.8 & 27.4 & 62.5 & 59.4 & 56.9 & 49.5 & 43.0 & 45.6 & 42.3 & 39.8 & 34.5 & 29.8 \\
    Swin~\cite{zhang2022dino} & 41.2 & 40.2 & 37.9 & 35.4 & 32.2 & 64.8 & 63.0 & 60.2 & 56.8 & 52.0 & 46.2 & 43.6 & 41.0 & 38.3 & 34.6 \\
    ConvNeXt~\cite{liu2022convnet} & 43.0 & 41.6 & 40.0 & 37.7 & 34.8 & 65.6 & 64.1 & 62.5 & 60.0 & 55.7 & 46.8 & 45.8 & 43.9 & 40.8 & 37.9 \\
    \bottomrule
  \end{tabular}
  }
  \label{tab:backbone}
\end{table*}

\begin{table}[t]
\caption{\textbf{TIDE~\cite{bolya2020tide} $\Delta AP_{O}~(AP_{O})$ measure on noise ratio at 20$\%$.}
  We investigate the types of errors the model makes due to each noise component. Lower is better.
  }
  \centering
  \resizebox{1\textwidth}{!}{
  \begin{tabular}{@{}l|c|cccccc@{}}
    \toprule
    Backbone & AP$_{50}$ & Cls & Loc & Both & Dupe & Bkg & Miss  \\
    \midrule
    ResNet~\cite{ren2015faster}      & 43.3 & 4.0~(47.3) & 5.8~(49.1) & 0.9~(44.2) & 0.2~(43.5) & 7.2~(50.6) & 9.7~(53.0)
 \\
    DCNv2~\cite{dai2017deformable} & 43.0 & 3.8~(46.8) & 5.4~(48.5) & 0.8~(43.9) & 0.1~(43.2) & 9.2~(52.2) & 8.3~(51.3)
 \\
    Swin~\cite{liu2021swin} & 52.0 & 2.9~(54.9) & 7.1~(59.2) & 0.9~(52.9) & 0.2~(52.2) & 7.9~(60.0) & 8.4~(60.4)
 \\
    ConvNext~\cite{liu2022convnet} & 55.7 & 3.4~(59.1) & 7.0~(62.7) & 1.0~(56.7) & 0.2~(55.9) & 7.1~(62.8) & 7.2~(62.9)
 \\
    \bottomrule
  \end{tabular}
  }
  \vspace{1em}
  \vspace{-15pt}
  \label{tab:TIDE_backbone}
\end{table}

As shown in~\tabref{tab:backbone} and~\tabref{tab:TIDE_backbone}, we conducted experiments to analyze the impact of the UNA setup on different backbones: Resnet-50~\cite{he2016deep}, Deformable Convolution~\cite{dai2017deformable}, Swin Transformer~\cite{liu2021swin}, and ConvNext~\cite{liu2022convnet}. 
Each of methods has a distinct receptive field, and we try to understand how these characteristics influence the performance in noise label. 
All result is based on FasterRCNN. For a more detailed analysis using LRP~\cite{oksuz2018localization}, please refer to Section 13 in the supplementary materials.

\paragraph{Deformable convolution network~(DCNv2~\cite{zhu2019deformable})} learns the spatial sampling within its receptive field from the data during training. When encountering noise labels, this characteristic of DCNv2 has a negative impact on performance. 
From Bkg in TIDE, it appears that DCNv2 frequently makes errors by mistaking the background as foreground.

\paragraph{Swin transformer~\cite{liu2021swin}} has a hierarchical structure and window-based self-attention mechanism, allowing it to capture long-range dependency while maintaining computational efficiency. Swin transformer incorporates global context information, which can help mitigate the negative impact of noise annotations.
Swin transformer exhibits higher localization errors compared to convolutional methods. This trend is also observed in DINO in~\tabref{tab:TIDE-algorithm}, suggesting that the high localization error might be attributed to the inherent characteristics of the Transformer.
\paragraph{ConvNext~\cite{liu2022convnet}} is a model that combines techniques and advancements from existing backbones, such as inverted bottleneck from MobileNet~\cite{howard2017mobilenets}, replacing ReLU~\cite{agarap2018deep} with GELU~\cite{hendrycks2016gaussian}, and layer normalization from Swin Transformer~\cite{liu2021swin}, to enhance the performance of convolutional networks. By incorporating advanced techniques from various models, ConvNext seems to have been designed to not only enhance performance but also address the challenges posed by noisy labels.
It is interesting to note that the TIDE distribution of ConvNext exhibits a pattern closer to Swin Transformer than traditional CNN models. %

\subsection{Learning algorithm}
\begin{table}[ht!]
  \caption{\textbf{Performance variation in the UNA setting based on the learning algorithm.} SSL stands for semi-supervised learning, and LNL refers to learning with noisy labels. The numbers in red and blue indicate performance decrease and increase, respectively.}
  \centering
  \resizebox{0.9\textwidth}{!}{
  \begin{tabular}{@{}cccccc@{}}
    \toprule
    & & \multicolumn{4}{c}{Noise Ratio} \\
    \cmidrule(r){3-6}
    Algorithm & Approach & 5\% & 10\% & 15\% & 20\% \\
    \midrule
    Baseline~\cite{ren2015faster} & Vanilla FasterRCNN & 35.7 & 33.1 & 29.9 & 26.6 \\
    OA-MIL~\cite{liu2022robust} & LNL in detection & 34.2 \textcolor{red}{(-1.5)} & 32.1 \textcolor{red}{(-1.0)} & 28.9 \textcolor{red}{(-1.0)} & 24.7 \textcolor{red}{(-1.9)} \\
    SoftTeacher~\cite{xu2021end} & SSL in detection & 36.0 \textcolor{blue}{(+0.3)} & 33.8 \textcolor{blue}{(+0.7)} & 31.6 \textcolor{blue}{(+1.7)} & 29.0 \textcolor{blue}{(+2.4)} \\
    \bottomrule
  \end{tabular}
  }
  \vspace{-5pt}
  \label{tab:learningalgorithm}
\end{table}

We evaluated the performance using OA-MIL~\cite{liu2022robust}, which is, 	
to the best of our knowledge, the only publicly available algorithm specifically designed for robust learning from noisy labels in object detection.~\tabref{tab:learningalgorithm} shows the difference in performance when OA-MIL is applied versus when it is not. Since OA-MIL is primarily focused on correcting localization noise, it falls short in the UNA setup that encompasses various types of noise. This highlights the need for methodologies that can address comprehensive noise. Additionally, as can be observed in~\tabref{tab:learningalgorithm}, we conducted experiments with the assumption that the semi-supervised object detection method, SoftTeacher~\cite{xu2021end}, could tackle the issue of missing labels. Although the improvement was modest, we noticed that the performance gains became more significant as the noise level increased.
These results demonstrate that addressing comprehensive noise requires a different approach from the existing methods for Learning with Noisy Labels in detection. 
The experimental results on the DeepLesion dataset~\cite{yan2018deeplesion}, which is a medical domain dataset, are documented in the appendix.

\section{Discussion}

Our research revealed that the study of noise labels in object detection tasks has been limited and there are overlooked aspects in this area. 
To analyze this, we classified label noise types into four categories and proposed a new problem setting called UNA (Unified Noise Annotation) that integrates them. 
We systematically investigated the performance variations of the detector in relation to the intensity of UNA setup. Additionally, we examined the specific errors that each detector tends to make using TIDE~\cite{bolya2020tide} and LRP~\cite{oksuz2018localization}, providing insights into the predominant mistakes made by each detector.
Based on these approaches, we traced the development of existing detection papers and analyzed the relationship between performance improvement and robustness to noise annotation.

Our finding is that the development of most detection algorithms improves performance in the presence of noise annotation. 
Advancements in object detection methods have been focused on eliminating hyper-parameters that require tuning and removing priors such as anchors. These approaches seems to reduce the impact of noise labels and improve the robustness of the models. 
One of the most recent and notable methods in this direction is DEtection TRansformer~(DETR)~\cite{carion2020end}. 
DETR-based methods demonstrate greater robustness to noise labels. For more discussion about detection transformer, please refer our supplementary materials.

In practical applications of detection in industry, it is common to generate labels directly. 
It is often observed that real-world scenarios fall short of the validated benchmark datasets, as the definition of labels may be somewhat ambiguous or different annotators may generate data based on varying criteria. 
We expect that our proposed UNA setup will minimize this gap.

\subsection{Broader impact}
\vspace{-5pt}
Numerous companies attempting to apply object detection into their industry struggle due to noisy labels. They are spending substantial amount of time to improve the quality of data label, especially in autonomous-driving or medical industries, where lives are at stake. By proposing a benchmark which is more practical compared to previous LNL for object detection and a framework which allows the detector to train robustly, we hope that LNL for object detection can be widely used in real-world situation.

\vspace{-5pt}
\section{Conclusion}
\vspace{-5pt}
In this paper, we analyzed various types of annotation noise for object detection that occur during dataset generation and propose a new benchmark called Unified Noise Annotation (UNA) to systematically analyze them. Through the systematic analysis using UNA, we observe the evolution of existing detection algorithms in terms of robustness to noise labels. Our findings highlight the importance of addressing annotation noise and its impact on the development of robust detection algorithms.

\begin{ack}
We would like to express our gratitude to Daeun Kyung for delivering an oral presentation on this research at the Medical Imaging meets NeurIPS 2023 workshop.
\end{ack}

\medskip
{\small
\bibliographystyle{plain}
\bibliography{egbib}
}

\end{document}